\documentclass{svproc}

\usepackage{type1cm}        
\usepackage{makeidx}         
\usepackage{graphicx} 
\usepackage{multicol}        
\usepackage[bottom]{footmisc}
\usepackage{subcaption} 

\usepackage{newtxtext}       %
\usepackage[varvw]{newtxmath}       

\makeindex             

\usepackage{todonotes}

\begin{document}

\mainmatter
\title{Stochastic Nonlinear Ensemble Modeling and Control for Robot Team Environmental Monitoring}
\titlerunning{Stochastic Nonlinear Ensemble Modeling and Control}
\author{Victoria Edwards \and Thales C. Silva \and M. Ani Hsieh}
\authorrunning{Victoria Edwards et al.}
\institute{University of Pennsylvania,
Department of Mechanical Engineering and Applied Mechanics, 
Philadelphia, USA,\\
\email{vmedw@seas.upenn.edu}}
%
%
\maketitle

\begin{abstract}
We seek methods to model, control, and analyze robot teams performing environmental monitoring tasks.
During environmental monitoring, the goal is to have teams of robots collect various data throughout a fixed region for extended periods of time. 
Standard bottom-up task assignment methods do not scale as the number of robots and task locations increases and require computationally expensive replanning. 
Alternatively, top-down methods have been used to combat computational complexity, but most have been limited to the analysis of methods which focus on transition times between tasks. 
In this work, we study a class of nonlinear macroscopic models which we use to control a time-varying distribution of robots performing different tasks throughout an environment. 
Our proposed ensemble model and control maintains desired time-varying populations of robots by leveraging naturally occurring interactions between robots performing tasks.
We validate our approach at multiple fidelity levels including experimental results, suggesting the effectiveness of our approach to perform environmental monitoring. 
\end{abstract}

\section{Introduction}
Robot teams performing long-term environmental monitoring require methods to manage different data collection across the team's varied sensing modalities.  
Since environments can be dynamic and uncertain, it is often necessary to vary the distribution of robots within the workspace over time as they perform their monitoring tasks. 
Existing strategies are mostly bottom-up where the objective is to synthesize individual robot strategies, \cite{khamis2015multi}, but are difficult to ensure the desired global distribution of the team over time.  
We present an approach to model and control time-varying distributions of robots with multi-modal sensing capabilities that extends existing top-down methods, \cite{Elamvazhuthi_2019}, by modeling robot interactions that govern changes in sensing behavior. 

To illustrate the importance of our approach, consider the long-term deployment of a robot team to monitor marine life in a littoral ocean environment. 
In these settings, autonomous surface vehicles (ASVs) may be tasked to measure the water depth and take images of the ocean floor but have limited storage capacity. 
While the ASVs can simultaneously collect both types of data, the limited onboard storage capacity may render it undesirable to collect image data during high tide when ocean floor visibility is reduced. 
Instead, at high tide some robots should focus on collecting water depth data while others focus on collecting images, and during low tide more should switch to collecting both depth and image data.  
This problem can be thought of as a planning and coverage problem, which couples the navigation of the robot to where the robots should sense \cite{almadhoun2019survey}. 

One approach to the problem of managing teams of robots with multiple capabilities is to model, plan, and control each individual robot. 
Microscopic modeling of teams of robots considers each robot and assigns robots to tasks as some variant of the resource allocation problem \cite{gerkey2003}.
Many algorithms exist to solve the task allocation problem \cite{khamis2015multi}. 
These methods often rely on heuristics to measure how well-suited a robot is for a specific task, and as such, they do not scale as the number of robots, number of sensing capabilities, and number of task locations increase.  
Furthermore, time-varying microscopic solutions require computationally expensive replanning \cite{Nam17}, and limited analytical tools exist to study the global properties of the team.  
As a result of these limitations, microscopic modeling restricts our ability to achieve the desired time-varying distributions of robots to collect the desired data.

Alternatively, macroscopic models represent global team behaviors and are most commonly derived through some mean field approximation with the assumption that the microscopic system is fundamentally stochastic \cite{Lerman05review,hsieh2008biologically}.
Macroscopic models have been employed to study and control the steady-state behavior of both homogeneous \cite{berman2009}, and heterogeneous cooperative swarms \cite{Hsieh09}.  
These models have also been used to synthesize control strategies to reduce the uncertainty in the microscopic systems \cite{Deshmukh2018,mather2011}.  
This is accomplished in \cite{mather2011} by explicitly modeling the first and second order moments of the team distribution in the workspace. 
Since macroscopic models are agnostic to team size and scale in the number of tasks, these approaches have been employed for homogeneous as well as heterogeneous robot teams \cite{prorok2017impact,ravichandar2020}. 
Nevertheless, these top-down approaches, similar to their bottom-up counterparts, require replanning when there is a new desired robot distribution because the macroscopic models employed are fundamentally linear. 
Various replanning methods for linear ensemble models have been proposed \cite{prorok2017impact,lee2019adaptive}, however these linear mean field models are fundamentally limited in terms of the complexity of microscopic behaviors that they can accurately represent  \cite{Elamvazhuthi_2019,harwell2021characterizing}.

Different from recent work \cite{biswal2021decentralized}, we propose a nonlinear ensemble model that can describe collaborative robot behaviors at the microscopic level and model the dynamics of time-varying robot distributions in the workspace. The proposed class of nonlinear ensemble models can explicitly account for robot-robot interactions and opens up new avenues for analyzing robot team behaviors. 
In this work, \textit{collaboration} during an environmental monitoring task means that robots probe the same event in space simultaneously in a distributed fashion. Since robots can also change their behaviors according to the overall environmental conditions, {\it e.g.}, tidal changes, {\it collaboration} refers to the robot's ability to adjust its short term goal to attain a global objective.  
To achieve a representation that exhibits those complex behavioral modes, we model the macroscopic continuous dynamics of the team using the Replicator Equations, which are commonly used in Biology to study natural selection and evolutionary dynamics \cite{sigmund1986survey}.
Alternative models have been considered to understand evolutionary dynamics with potential applications to robotics, but existing work primarily focus on the continuous analysis of discrete solutions overlooking the fundamentally stochastic nature of the microscopic multirobot system \cite{leonard2014multi,pais2012hopf,dey2018feedback}.
Instead, we use the Replicator Equations to model a population of robots interacting in a collaborative fashion, where the collaboration necessitates a time-varying distribution of robots across possible behavioral modes and/or specific regions in the workspace.

The contributions in this paper are: 1) the development of a nonlinear macroscopic model for a collaborative robot team based on the Replicator Equations, 2) the design of a feedback control strategy to handle uncertainties during team deployment based on the proposed nonlinear macroscopic model, and 3) a method to distribute macroscopic model derived feedback control strategy for the team. 
We show how our proposed framework works well for environmental monitoring tasks that requires robots with several sensing capabilities to work collaboratively over time through our multi-level fidelity simulation and experimental results.

\section{Problem Formulation}
We are interested in problems that require robot teams to exhibit time-varying behavior distributions.
Unlike existing linear macroscopic ensemble methods which consider transition times \cite{Elamvazhuthi_2019}, our approach uses collaboration rates, a measure of the interaction between robots that share spatial proximity, in a class of nonlinear models. 
We illustrate our method with two example time-varying solutions which are not possible with existing linear macroscopic methods. 


\subsection{Task topology}
\label{sec:robot_model}
Consider a set of $M$ tasks where for each task a robot must exhibit a specific set of behaviors, for example sensing a specific phenomenon. 
We build an undirected graph to represent the tasks and potential switching between task.
The graph $\mathcal{G} = (\mathcal{V}, \mathcal{E})$, has nodes for each task $v_j \in \mathcal{V}$ where $j = 1, ..., M$, and edges $e_{ij} \in \mathcal{E}$.
If an edge $e_{ij} = 0$ then it is not possible for robots to switch from task $i$ to task $j$, likewise, when $e_{ij} = 1$ it is possible for robots to switch from task $i$ to task $j$. 
We assume the graph $\mathcal{G}$ is strongly connected, which means that there is a path from each task to any other task in the graph. 
Note that the existence of an edge is not reflective of distance to a spatially distributed task as in \cite{mather2011,prorok2017impact}, but instead reflective of the potential for robot collaboration to occur. 

\subsection{Ensemble model}
\label{sec:model}
Given $N$ robots and $M$ collaborative tasks, let $Y_i$ be a random variable which represents the population of robots at task $i$, and the vector of random variables for all tasks be $Y = [ Y_1 ~ ... ~ Y_M ]^T$.
We present a class of models that describe the stochastic jump process for how the robot ensemble evolves in time.
Consider two collaborating robot populations, where robots performing task $i$ and robots performing task $j$ are expected to have spatially close interactions, then
\begin{align}
    Y_i + Y_j \xrightarrow[]{k_{ij}} 2 Y_i,
    \label{eq:stochastic_jump_process}
\end{align}
where $k_{ij}$ describes the Poisson collaboration rate between the two tasks $i$ and $j$ that will exist for every nonzero edge in the task graph $\mathcal{G}$.
The physical intuition for a nonzero collaboration rate at the microscopic level is when two agents performing different tasks have physically close interactions with one another
the result will be one agent changing task.
Each collaboration rate $k_{ij}$ can be put into a matrix $K \in \mathbb{R}^{M\times M}$, commonly called the payoff matrix. 
An entry of the payoff matrix, $k_{ij}$, maps 
what will happen when an interaction happens between two agents performing task $i$ and task $j$. 
Consider a row $i$ of the payoff matrix, anywhere that $k_{ij} = 0$, means that when a robot performing task $i$ interacts with a robot performing task $j$ nothing happens. 
We abuse notation by letting the signal of $k_{ij}$ be negative or positive to express two types of switches. 
Specifically, if $k_{ij} > 0$ then the process results in a robot changing from task $Y_j$ and starting to perform task $Y_i$, whereas
if $k_{ij}< 0$ then the process results in a robot changing from task $Y_i$ and starting to perform task $Y_j$.

The Replicator Equations \cite{sigmund1986survey}, are the dynamical systems which use the provided collaboration rates to describe the evolution of different agent behaviors and can be written in vector form as, 
\begin{align}
  \dot{Y}_i = Y_i [ \sum_j^M k_{ij} Y_j - \sum_p^M \sum_l^M Y_p k_{pl} Y_l].
  \label{eq:rep_general}
\end{align}
The Replicator Equations in \eqref{eq:rep_general} can be constrained to satisfy the assumption that $\sum_i^M Y_i = 1$, such that, if the initial state $Y(0)$ satisfies this constraint, then all future states will satisfy this condition--maintaining a fixed population of robots $N$ \cite{sigmund1986survey}. 
Note that Equation \eqref{eq:stochastic_jump_process} only considers collaborations between two task types, while there are third order terms accounted for in the Replicator Equations.
This simplification is justified because it is of low likelihood to have three robots of the appropriate type collaborating.

This formulation allows us to consider the collaborations between robot populations and study how spatial robot-robot interactions change the ensemble dynamics. 
The Replicator Equations in \eqref{eq:rep_general} will behave differently for different structures of the payoff matrix, $K$, because each entry can be $0$ or nonzero depending on the allowed collaborations  specified by $\mathcal{G}$. 
We will consider two relevant examples throughout the paper which have equilibrium that guarantee time-varying distributions of robots that could be used in the littoral ocean monitoring scenario. 


\textbf{Example 1:} Consider the canonical example of the Lokta Volterra model, written as the Replicator Equations in \eqref{eq:rep_general} as detailed by Hofbauer \cite{hofbauer1981}. 
This example has three tasks, $M = 3$, and the payoff matrix is
\begin{align}
K = \begin{bmatrix} 
0 & 0 & 0 \\ k_{10} & 0 & -k_{12} \\ -k_{20} & k_{21} & 0\end{bmatrix}.
\end{align} 
The rules from potential interactions to collaboration are 
\begin{align}
    Y_0 + Y_1 \xrightarrow[]{k_{10}} 2 Y_1, \\ \nonumber
    Y_1 + Y_2 \xrightarrow[]{k_{12}} 2 Y_2, \\ \nonumber
    Y_0 + Y_2 \xrightarrow[]{k_{20}} 2 Y_0, \\ \nonumber
    Y_1 + Y_2 \xrightarrow[]{k_{21}} 2 Y_2, \\ \nonumber
\end{align}
which correspond to the following differential equations that describe how the processes evolve in time, 
\begin{align} \nonumber
  \dot{Y}_0 & = Y_0 (k_{20}Y_0Y_2 - k_{10} Y_0Y_1 + (k_{12} - k_{21})Y_1Y_2), \\ \nonumber
  \dot{Y}_1 & = Y_1(k_{10}Y_0 - k_{12}Y_2 + k_{20}Y_0Y_2 - k_{10}Y_0Y_1 + (k_{12} - k_{21})Y_1Y_2),  \\ 
  \dot{Y}_2 & = Y_2(k_{21} Y_1 - k_{20}Y_0 + k_{20}Y_0Y_2 - k_{10}Y_0Y_1 + (k_{12} - k_{21})Y_1Y_2). 
  \label{eq:example1}
\end{align}
The parameters $k_{10}$, $k_{12}$, $k_{20}$, and $k_{21}$, are the collaboration rates, and define the result of an interaction between two robots performing task $i$ or task $j$.
The possible outcomes based on the collaboration rate are; keep doing the the same tasks, or have one of the robots change task. 

This model is of interest to us because it exhibits time-varying population distributions.
Consider the following equilibrium for positive robot populations, $Y^d = \begin{bmatrix} 0, & \frac{k_{12}}{k_{12} - k_{21}}, & \frac{k_{21}}{k_{21} - k_{12}} \end{bmatrix}^T$.
If we linearize Equation \eqref{eq:example1} about $Y^d$, and compute the eigenvalues of the linearization we get $\lambda_0 = 0$,
$\lambda_1 = -\frac{i k_{21}\sqrt{k_{10}}}{\sqrt{k_{20}}}$, and $ \lambda_2 =  \frac{i k_{21}\sqrt{k_{10}}}{\sqrt{k_{20}}}$. 
From dynamical systems theory we know that the system is classified as a center. (See Guckenheimer and Holmes Ch 1 \cite{guckenheimer2013} for a detailed
analysis.) 
This means that for the same set of parameters, $K$, and different initial positions, $Y(0)$, a unique non-intersecting orbit will occur around the fixed point. 
The oscillations that result from a center equilibrium produce time-varying distributions of robots.
From our littoral ocean monitoring case, we can select parameters which match the tidal charts so that the desired sensing measurements are performed for representative amounts of time without wasting space resources with unnecessary data or compute resources for replanning.  

\textbf{Example 2}: Consider a four task example, $M = 4$, discussed in detail by Schuster et al. \cite{schuster1981selfregulation}. 
The payoff matrix is written as follows, 
\begin{align}
K = \begin{bmatrix}0 & 1 & -\mu & 0 \\ 0 & 0 & 1 & -\mu \\ -\mu & 0 & 0 & 1 \\ 1 & -\mu & 0 & 0 \end{bmatrix}.
\label{eq:payoff_rep4d}
\end{align}
The collaboration definitions can be easily achieved by using Equation \eqref{eq:stochastic_jump_process}, and the resulting dynamical system using Equation \eqref{eq:rep_general} comes out to,
\begin{align}\nonumber
\dot{Y}_1& = Y_1 (Y_2 - \mu Y_3 + 2 \mu Y_1 Y_3 + 2 Y_2 Y_4 - Y_1Y_2 - Y_2Y_3 - Y_4Y_3 - Y_4 Y_1),\\\nonumber
    \dot{Y}_2 &= Y_2 (Y_3 - \mu Y_4 + 2 \mu Y_1 Y_3 + 2 Y_2 Y_4 - Y_1Y_2 - Y_2Y_3 - Y_4Y_3 - Y_4 Y_1),\\\nonumber
    \dot{Y}_3 & = Y_3 (Y_4 - \mu Y_1+ 2 \mu Y_1 Y_3 + 2 Y_2 Y_4 - Y_1Y_2 - Y_2Y_3 - Y_4Y_3 - Y_4 Y_1),\\
    \dot{Y}_4 & = Y_4 (Y_1 - \mu Y_2+ 2 \mu Y_1 Y_3 + 2 Y_2 Y_4 - Y_1Y_2 - Y_2Y_3 - Y_4Y_3 - Y_4 Y_1). 
    \label{eq:example2}
\end{align}
Notice that for $k_{ij} = 1$ an interaction results in a change from task $j$ to task $i$, and in the case that $k_{ij} = -\mu$ the result will be a change from task $i$ to task $j$.

Again this model is of interest because of equilibrium behavior which provides alternative time-varying population distributions. 
Shuster et al. proposed the following equilibrium $Y^d = \begin{bmatrix} \frac{1}{4}, & \frac{1}{4}, & \frac{1}{4}, & \frac{1}{4} \end{bmatrix}^T$, and if we linearize Equation \eqref{eq:example2} about $Y^d$,
and compute the eigenvalues of the linearization we get $\lambda_0 = \frac{-1 + \mu}{4}$, $\lambda_1 = \frac{-i + \mu}{4} $, $\lambda_2 = \frac{i + \mu}{4}$, and $\lambda_3 = \frac{- 1 - \mu}{4}$.
The analysis that Schuster et al. performed concluded that at the equilibrium point $Y^d$, $\mu$ is a bifurcation parameter. 
When $\mu < 0$ the system is stable because the real part of all eigenvalues are negative. 
At $\mu = 0$ the system is at the bifurcation point, where the complex plane is crossed and the negative real parts of $\lambda_2$ and $\lambda_3$ vanish. 
Finally, when $\mu > 0$ the system exhibits a stable Hopf Bifurcation for small enough $\mu$. 
(See Schuster et al. \cite{schuster1981selfregulation} Section 4 for further details on the analysis). 
The existence of a stable Hopf Bifurcation means that an attracting limit cycle exists in some neighborhood of the equilibrium. 
This means that the resulting robot populations will fluctuate at regulated intervals governed by the limit cycle. 
In the littoral ocean monitoring task, we can break down the tides into four periods of interest and pick parameters to match such that different sensing happens based on local interactions during the corresponding time with minimal replanning or information exchanged.

\section{Ensemble Model Control}
Our goal is to maintain proximity to an equilibrium of interest which provide time fluctuating distributions of robots performing different behaviors. 
Previous work has proposed controlling different aspects of the linear mean field model by considering different nonlinear feedback strategies \cite{mather2011,Deshmukh2018}. 
We will aim to maintain a trajectory by actuating on the collaboration rates to control how the interaction with other agents influences the switch to different behaviors.
The macroscopic continuous solutions to our proposed models give deterministic behavior, and will result in constant oscillations either around the center equilibrium in Example 1 or the limit cycle equilibrium in Example 2. 
However, because our models are representations of stochastic processes we know that due to variability in the robot-robot and robot-environment interactions our desired distributions may not be achieved over time. 
This is important because if the environment we are monitoring in has specific windows where phenomena can be observed it is necessary that we stay near the desired distributions. 

\subsection{Feedback Controller}
We are proposing a modification to the collaboration rate, $k_{ij}$, to take as feedback a desired trajectory, $Y_i^*(t)$. 
The evaluation for $Y_i^*$ is achieved from the numerical solution of 
the continuous macroscopic differential equations, Equation \eqref{eq:rep_general}, with the desired parameters. 
Consider the following trajectory tracking controller,
\begin{align}
    k_{ij} = \alpha_{ij}(\frac{Y_i^*(t)}{Y_i(t)} - 1),
    \label{eq:control}
\end{align}
where $\alpha_{ij}$ is a tunable parameter, $Y^*_i(t)$ is the desired trajectory and $Y_i(t)$ is the current population fraction at time $t$. 
If we plug this new collaboration rate into Equation \eqref{eq:stochastic_jump_process} we get: 
\begin{align}
    Y_i + Y_j \xrightarrow[]{\alpha_{ij}(\frac{Y_i^*}{Y_i} - 1)} 2 Y_i.
\end{align}
After substituting Equation \eqref{eq:control} into the Replicator Equations in \eqref{eq:rep_general} we get: 
\begin{align}\nonumber
      \dot{Y_i} & = Y_j [ \sum_j^M (\alpha_{ij}(\frac{Y_i^*}{Y_i} - 1)) Y_i - \sum_p^M \sum_l^M Y_l \alpha_{pl}(\frac{Y_p^*}{Y_p} - 1) Y_p], 
      \\
      & = Y_j [\sum_j^M\alpha_{ij}(Y_i^* - Y_i) - \sum_p^M \sum_l^M \alpha_{pl}Y_l(Y_p^* - Y_p)].
\end{align}
The resulting $(Y_i^* - Y_i)$ acts as a proportional error term which regulates the population based on the desired trajectory.
Notice that, as before, we will have collaboration rates that are negative which will switch the resulting collaboration outcome.
The same substitutions can be replicated for both Example 1 and Example 2. 

\subsection{Distributed Microscopic Algorithm}
To demonstrate that the proposed nonlinear stochastic representations are beneficial for modeling and controlling robot teams we will outline an idealized microscopic scenario, which acts as a proxy for teams performing environmental monitoring.
The goal is to find the area of interaction, $a_{ij}$ and corresponding radius of interaction, $r_{ij}$, which are microscopic instantiations of the macroscopic collaboration rate $k_{ij}$.
If there is a nonzero collaboration rate, $k_{ij} \neq 0$, then we know that when robots interact within a certain area a task switch will occur.
Notice that if the collaboration rate is zero, than the robots may interact but this interaction will not result in collaboration. 
The area of interaction is directly related to the collaboration rate, which is a measure of how close two agents need to be to one another for a task switch to happen.
Let $\hat{Y_i}$ be the population count for robots performing task $i$, then the area of interaction, $a_{ij}$, is computed as follows
\begin{align}
a_{ij} = \frac{A k_{ij}}{v_{ij}\hat{Y_i}\hat{Y_j}},
\end{align}
where $A$ is the area of the environment to monitor, $k_{ij}$ is the collaboration rate, $v_{ij}$ is the relative speed of robot $i$ and robot $j$, and $\hat{Y}_i \hat{Y}_j$ are all disjoint pairs of collaborations possible given the populations of robots performing these tasks.  
The robot radius of interaction is: $r_{ij} =  \sqrt{a_{ij}/\pi} $.

Incorporating the proposed feedback controller, Equation \eqref{eq:control}, consider 
\begin{align}
    a_{ij} = \frac{A k_{ij}}{v_{ij} \hat{Y}_j\frac{1}{\hat{Y}_i^* - \hat{Y}_i}},
    \label{eq:interaction_area}
\end{align}
where $\hat{Y}_i^* = Y_i^*N$ is the re-scaling of the desired trajectory population fraction by the total number of robots in the system.
The intuition for the feedback term is that when $\hat{Y}_i$ is far from $\hat{Y}_i^*$ then the resulting 
area of interaction will increase.
Likewise, when $\hat{Y}_i$ is close to $\hat{Y}_i^*$ the area of interaction decreases.
The interaction area is updated in time as the microscopic simulations are running. 

We extend our controller to consider distributed estimation of the population dynamics.
Similar to other methods, \cite{lee2019adaptive,biswal2021decentralized}, we are proposing a local estimation method of robots performing tasks.  
Consider that each robot has a fixed sensing radius, $\mathcal{R}_i$, and the ability to determine which robots within its radius are performing which behaviors. 
The neighborhood of the robot, $\mathcal{N}_i$, is every robot within $\mathcal{R}_i$, and the population estimate for robots performing different $i$ tasks within $\mathcal{N}_i$ is $\tilde{Y_i}$.
This method assumes that within the sensing radius, $\mathcal{R}_i$, there will be a sufficient density of robots to act as a proxy for the overall distributions of robot behaviors.
Substituting $\tilde{Y}_i$ for $\hat{Y}_i$ in Equation \eqref{eq:interaction_area} gives a new locally computed interaction area.
Note to scale the desired trajectory, $\tilde{Y}_i^* = Y_i^* \mathcal{N}_i$.
Using local estimates and an appropriately scaled desired trajectory we have eliminated the need for global information in our microscopic experiments.

\section{Simulation and Experimental Setup}
We used three levels of fidelity to evaluate our proposed approach: macro-discrete simulation trials, microscopic simulation trials, and mixed reality experimental trials. Parameters for all results were selected to be near our equilibria of interest for both Example 1 and Example 2, which ensure that we see time-varying distributions of robots performing tasks. 

\begin{figure}[ht]
  \centering
    \centering
    \includegraphics[width=\textwidth]{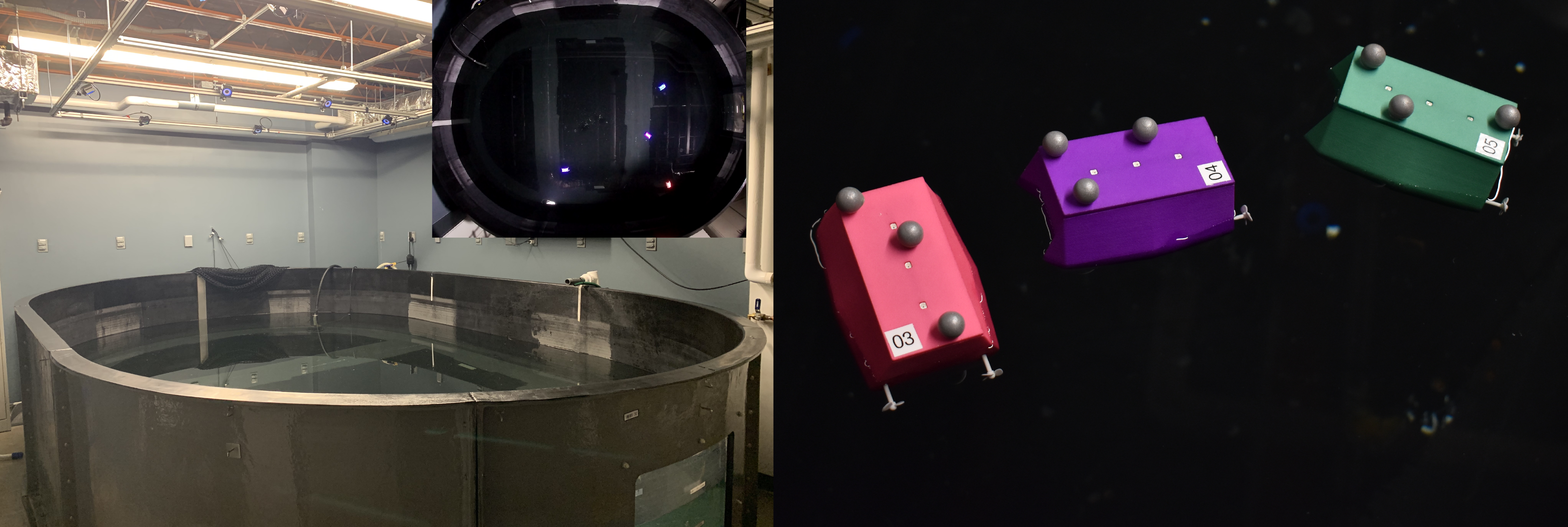}
    \caption{The multi-robot Coherent Structure Testbed (mCoSTe) water tank where the miniature Autonomous Surface Vehicles (mASV) perform different tasks.  }
    \label{fig:exp_setup}
\end{figure}

Our macro-discrete simulations used the Stochastic Simulation Algorithm (SSA) from Gillespie \cite{gillespie1977exact}. 
SSA takes in initial populations of robots $\hat{Y}(0)$ and collaboration rates $K$.
To determine the next time and type of event, SSA uses a pseudo random number generator. 
While the total time and total number of transitions are not exceeded, the following two steps are iteratively taken.  
The next event is determined, by considering the likelihood of a certain task switch based on the current population counts, $\hat{Y}_i$ and the collaboration rates, $k_{ij}$. 
Consider that each collaboration type has the following scaled collaboration rate,
\begin{align}
b_{ij} = k_{ij}\hat{Y}_i \hat{Y}_j.
\end{align}
Select a random number, $r_1$ and scale that value by $b_n = \sum_i^M \sum_j^M b_{ij}$, or the total of all possible collaborations.
The next event is which ever, $b_{ij}$ is closest to the scaled random number.
Note that the resulting switch is defined by reactions outlined in Equation \eqref{eq:stochastic_jump_process}, and the population counts $\hat{Y}_i$ and $\hat{Y}_j$ are updated accordingly.
Next, the time, $t_p$, is updated to the next time, $t_{p+1}$, by generating a random number $r_2$ such that,
\begin{align}
    t_{p+1} = t_p + \log{\bigg(\frac{1}{r_2}\bigg)} \frac{1}{b_n}.
\end{align}
The process continues until the maximum time or maximum number of transitions are met. 
An important observation is that SSA ignores robot dynamics, and considers only the stochastic jump process making it mathematically equivalent to microscopic simulations.

Specific to our implementation, the original Replicator Equations in \eqref{eq:rep_general}, require initial conditions to sum to $1$, which is not compatible with SSA.
To deal with this we scale the population fractions by $N$, the result of doing this is a shift in the magnitude of the parameters used for the macroscopic continuous solutions and macro-discrete trials. 
In addition, to prevent behavior extinction, we enforce that each behavior is performed by at least one agent at any given time. 

To further validate our method, microscopic simulation trials were introduced to incorporate robot dynamics. 
Robots were simulated using ideal point particles, have unit velocity, and collide with the boundary of the workspace using an angle of deflection of $\pi/2$. 
A difference from the macro-discrete trials is that robots now compute a direct interaction radius $r_{ij}$, based on the collaboration rates. 
This means when an encounter with another robot occurs, if there is a nonzero collaboration rate and agents are spatially within $r_{ij}$, then the predetermined interaction task switch ensues. 
Trials were run for $100$ seconds, with a fixed number of robots represented by point particles in an $18 \times 18$ unit area. 

Experimental results were performed in the multi-robot Coherent Structure Testbed (mCoSTe), Figure \ref{fig:exp_setup}, which includes the miniature Autonomous Surface Vehicles (mASV), and a Multi Robot Tank (MR tank).
The MR Tank is $4.5 \mathrm{m} \times 3.0 \mathrm{m} \times 1.2 \mathrm{m}$ water tank equipped with an OptiTrack motion capture system. 
The mASVs are a differential drive platform, localize using 120 Hz motion capture data, communicate via XBee, and onboard processing is done with an Arduino Fio.
Mixed reality experiments were performed to increase the robot density, due to limited number of mASV platforms available. 
Experiments use 4 mASV and 6 idealized point particles with the same configuration as the microscopic simulation trials.
To implement our method on the mASV we assigned waypoints to robots by randomly selecting positions on the environmental boundary, which had an area of $2.3\mathrm{m} \times 1.3 \mathrm{m}$. 
During the mixed reality experiment, robots react to simulated agents as if they were in the environment.
Behavior switches occur based on the interaction radius $r_{ij}$.
This means when any agent, robot or simulated, is within $r_{ij}$ the predetermined collaboration switch happens. 
The interaction radius comes directly from the calculation of the area of interaction, Equation \eqref{eq:interaction_area}, which is the connection between the macroscopic model and the microscopic implementation.

\section{Results}

\begin{figure}[ht]
\centering
    \begin{subfigure}{0.495\textwidth}
    \centering
    \includegraphics[width=\textwidth]{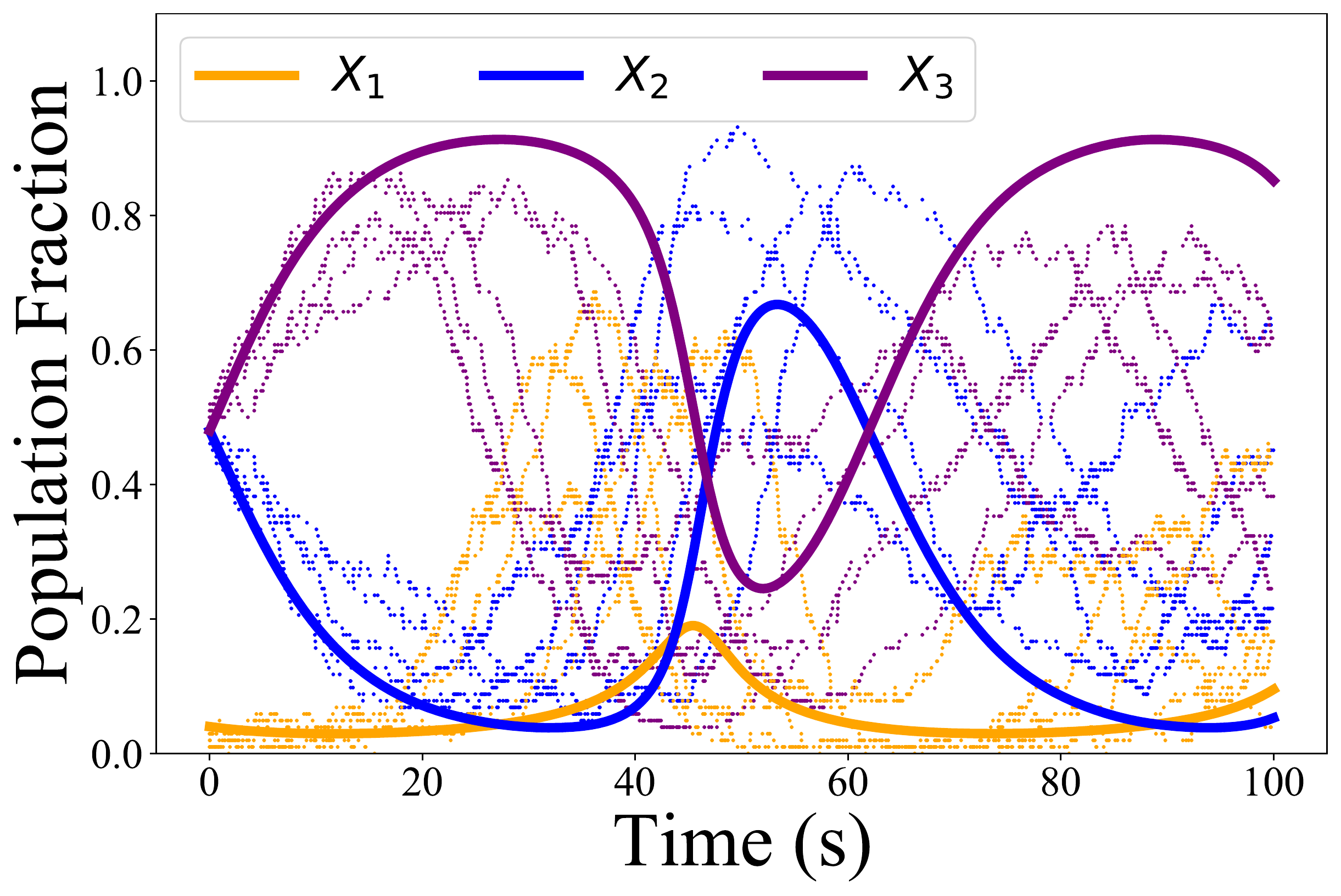}
    \caption{Without trajectory feedback}
    \label{fig:example_1_no_control}
    \end{subfigure}
    \begin{subfigure}{0.495\textwidth}
    \centering
    \includegraphics[width=\textwidth]{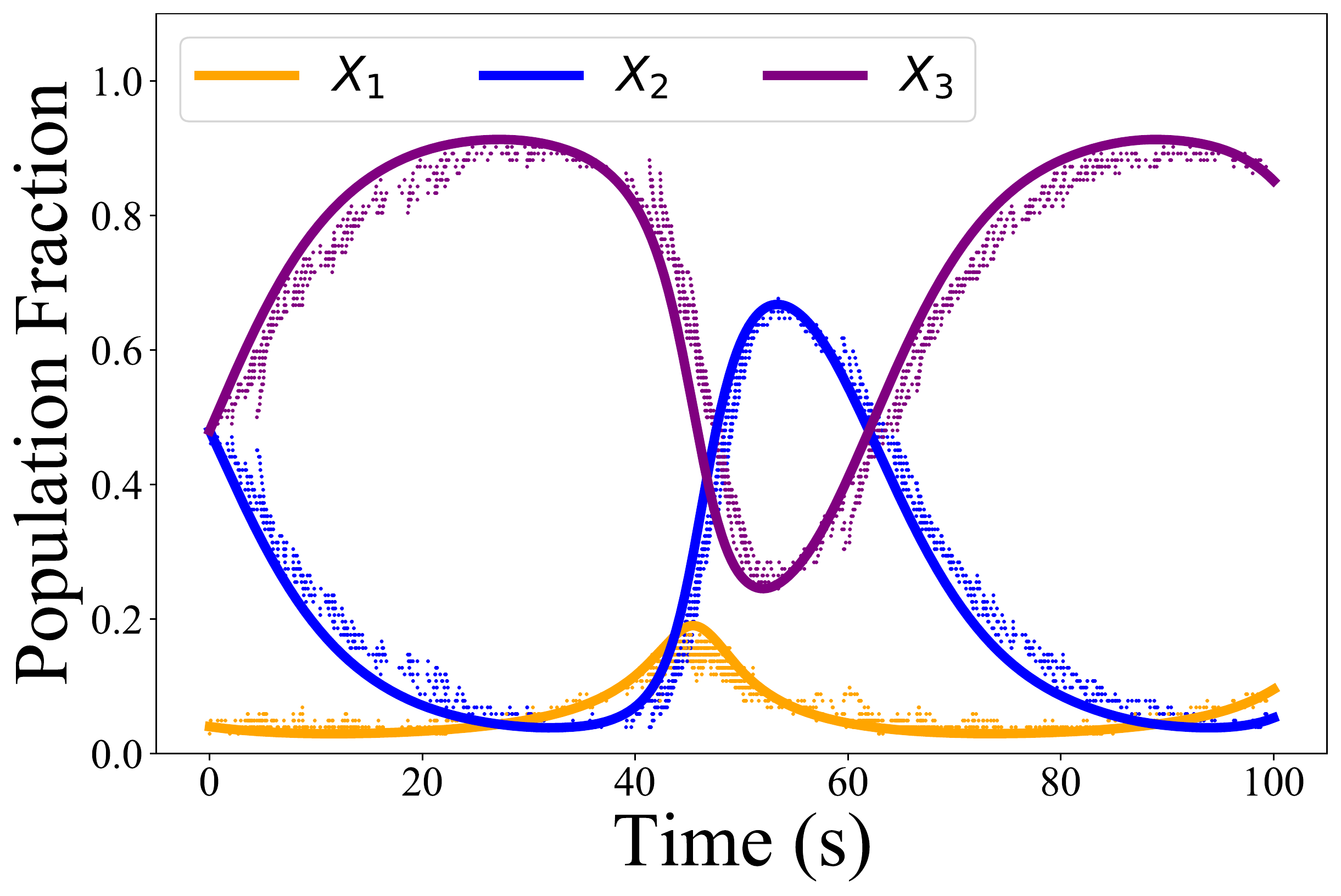}
    \caption{With trajectory feedback}
    \label{fig:example_1_control}
    \end{subfigure}
    \caption{Example 1 has three time-varying tasks, and include five SSA trials. Figure (a) parameters: $\hat{Y}(0) = \begin{bmatrix} 2 & 50 & 50 \end{bmatrix}^T$, with $k_{10} = 0.004$, $k_{12} = 0.0004$, $k_{20} =  0.003$, and $k_{21} = 0.0008$. Figure (b) parameters: $\hat{Y}(0) = \begin{bmatrix} 2 & 50 & 50 \end{bmatrix}^T$, with $\alpha_{10} = 0.03$, $\alpha_{12} = 0.003$, $\alpha_{20} =  0.0225$, and $\alpha_{21} = 0.006$.}
    \label{fig:example_1_macro_discrete_results}
\end{figure}

\begin{figure}[ht]
  \centering
    \begin{subfigure}{0.495\textwidth}
    \centering
    \includegraphics[width=\textwidth]{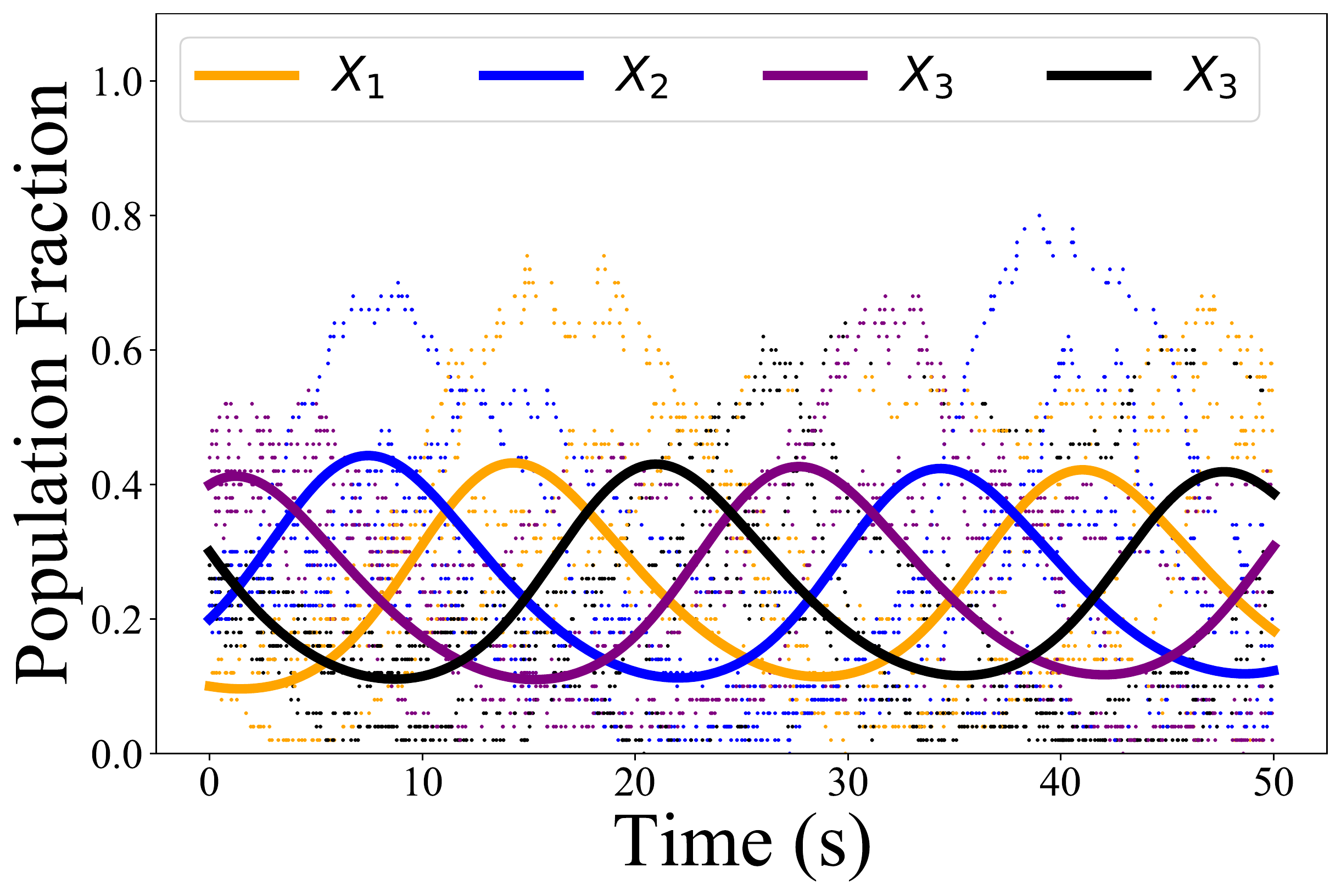}
    \caption{Without trajectory feedback}
    \label{fig:example_2_no_control}
    \end{subfigure}
    \begin{subfigure}{0.495\textwidth}
    \centering
    \includegraphics[width=\textwidth]{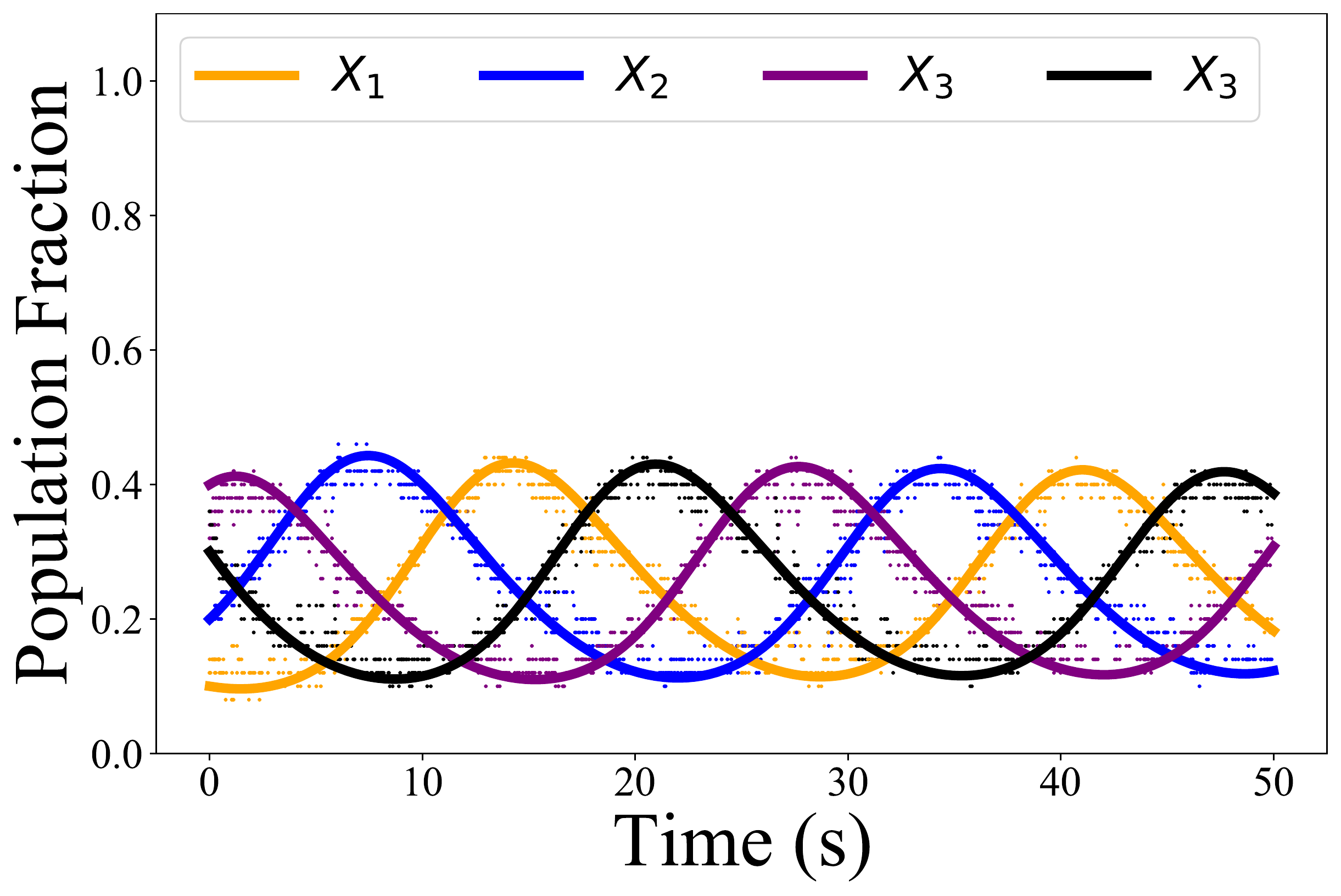}
    \caption{With trajectory feedback}
    \label{fig:example_2_control}
    \end{subfigure}
    \caption{ Example 2 has four time varying tasks and plots include five SSA trials. Figure (a) parameters: $\hat{Y}(0) = \begin{bmatrix} 5 & 10 & 15 & 20 \end{bmatrix}^T$, with $k_{12} = k_{23} = k_{34} = k_{41} = 0.01$ and $\mu = 0.0001$. Figure (b) parameters: $\hat{Y}(0) = \begin{bmatrix} 5 & 10 & 15 & 20 \end{bmatrix}^T$, with $\alpha_{12} = \alpha_{23} = \alpha_{34} = \alpha_{41} = 0.1$, and $\mu = 0.001$}
    \label{fig:example_2_macro_discrete_results}
\end{figure}

Our results use multiple levels of fidelity to demonstrate how interactions can induce time varying distributions of robot behaviors.
When possible the same initial conditions were used for both Example 1 and Example 2 to help draw comparisons between different types of results. 
Specifically, the desired trajectories for Example 1 and Example 2 were the same for all fidelity results. 
The initial conditions for Example 1 were a reference trajectory of $Y^*(0) = \begin{bmatrix} 0.2 & 0.2 & 0.6\end{bmatrix}^T$ with parameters $k_{10} = 2.0$, $k_{12} = 0.2$, $k_{20} =  1.5$, and $k_{21} = 0.4$. 
The initial conditions for Example 2 were a reference trajectory of $Y^*(0) = \begin{bmatrix} 0.1 & 0.2 & 0.4 & 0.3 \end{bmatrix}^T$ and parameter $\mu = 0.01$.
Figures contain reference trajectories $Y^*$ as solid lines, and resulting population fractions as dots. 
Further details for parameters used for each level of fidelity simulation and experiment are in the corresponding figure captions. 
For both Example 1 and Example 2 we selected parameters near the equilibrium presented in Section \ref{sec:model} so that time-varying distributions of robot behaviors were observed. 

Five macrodiscrete, SSA, trail results are in Figure \ref{fig:example_1_macro_discrete_results} and \ref{fig:example_2_macro_discrete_results}.
In the case without control, Figures \ref{fig:example_1_no_control} and \ref{fig:example_2_no_control}, each trial roughly follows the desired time varying trajectory, but the variability is significant. 
This variability can be attributed to microscopic behaviors not being accounted for during SSA.
In the case with control, Figures \ref{fig:example_1_control} and \ref{fig:example_2_control}, the trials follow the trajectory closely as we anticipated. 
The gains, $\alpha_{ij}$ were tuned so that the rate of interactions were sufficiently high, if there was an insufficient interaction rate then the SSA trials would lag behind the reference trajectory.  

\begin{figure}[ht]
  \centering
    \begin{subfigure}{0.495\textwidth}
    \centering
    \includegraphics[width=\textwidth]{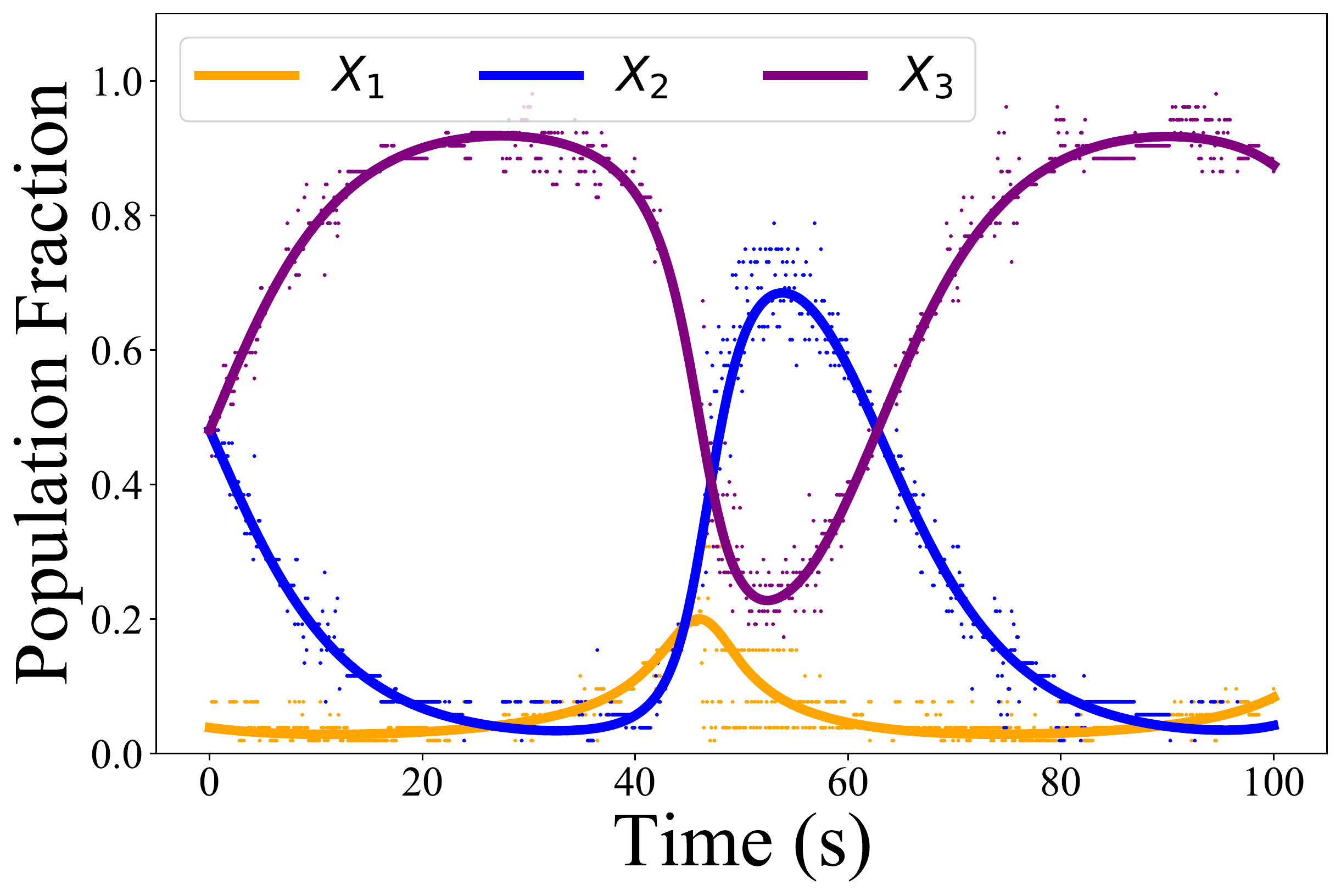}
    \caption{Centralized Microscopic Result}
    \label{fig:example_1_centralized_micro}
    \end{subfigure}
    \begin{subfigure}{0.495\textwidth}
    \centering
    \includegraphics[width=\textwidth]{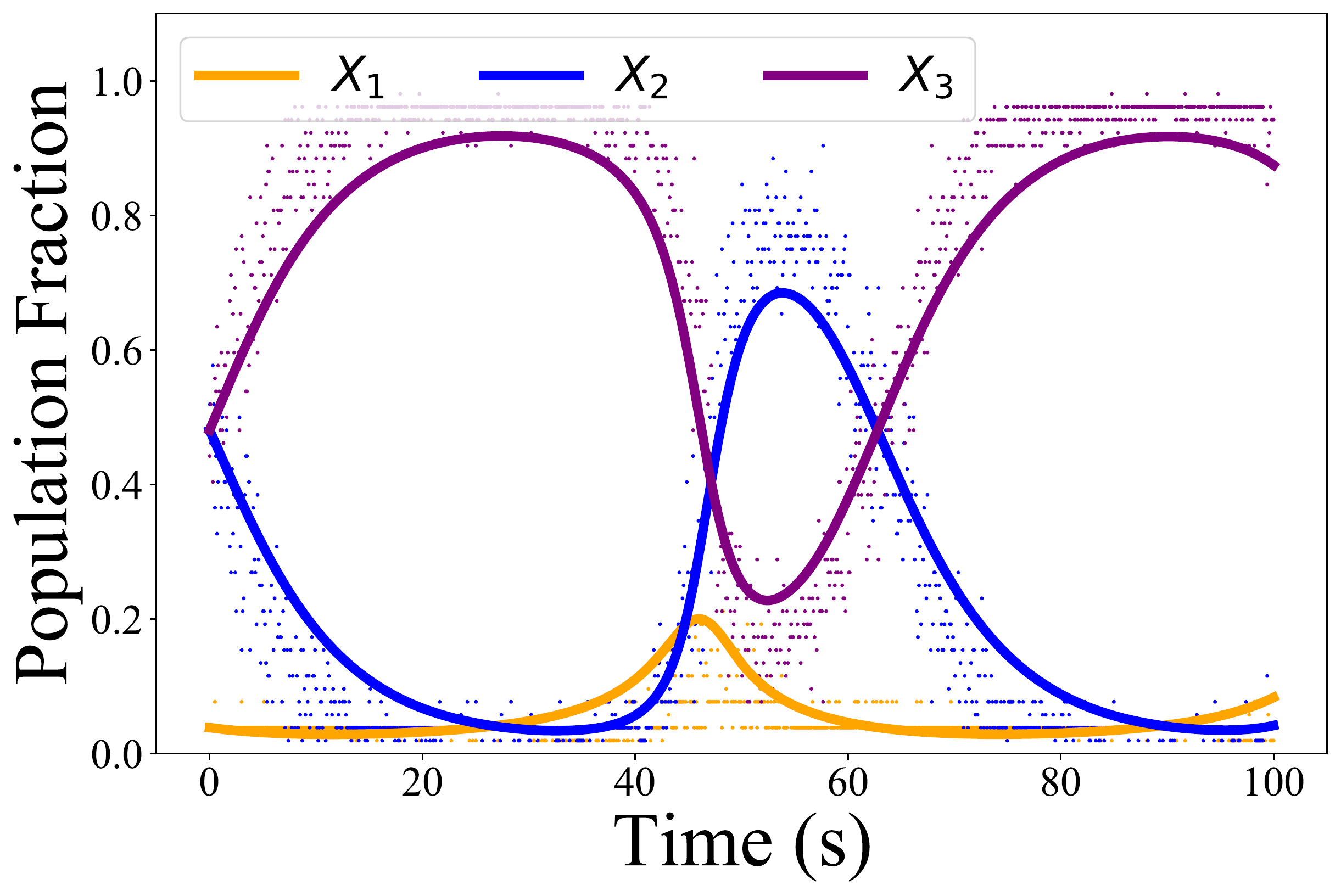}
    \caption{Distributed Microscopic Result}
    \label{fig:example_1_decentralized_micro}
    \end{subfigure}
    \caption{Controlled microscopic results of Example 1. Figure (a) uses global population estimation, and Figure (b) uses distributed population estimation where $\mathcal{R}_i = 5$. Both trials use $\hat{Y}(0) = \begin{bmatrix} 2 & 25 & 25\end{bmatrix}^T$, $\alpha_{10} = 2.0$, $\alpha_{12} = 0.2$, $\alpha_{20} =  1.5$, and $\alpha_{21} = 0.4$.}
    \label{fig:micro_validation_example_1}
\end{figure}

\begin{figure}[ht]
  \centering
    \begin{subfigure}{0.495\textwidth}
    \centering
    \includegraphics[width=\textwidth]{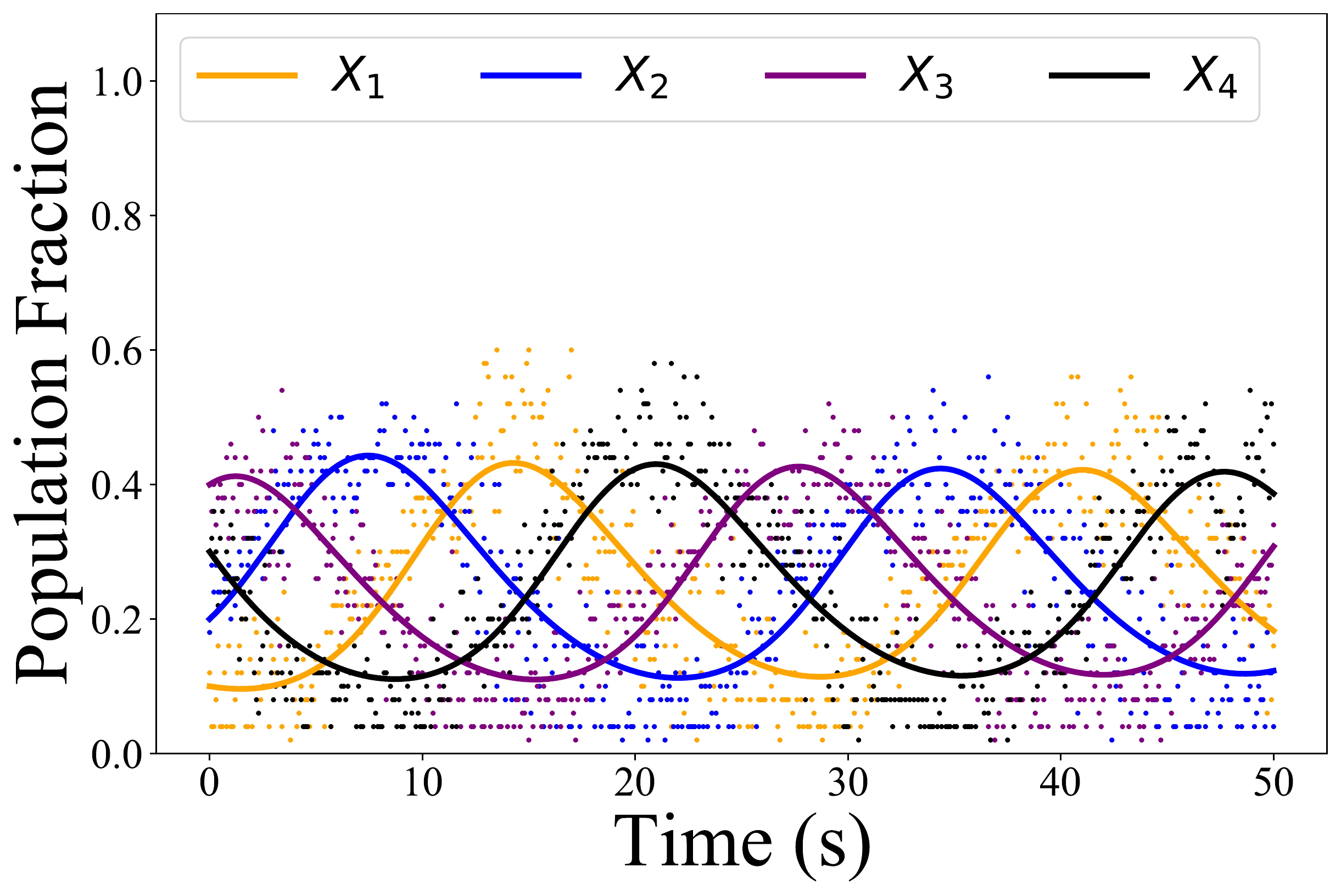}
    \caption{$\mathcal{R}_i = 10$ Microscopic Result}
    \label{fig:example_2_decentralized_micro_r10}
    \end{subfigure}
    \begin{subfigure}{0.495\textwidth}
    \centering
    \includegraphics[width=\textwidth]{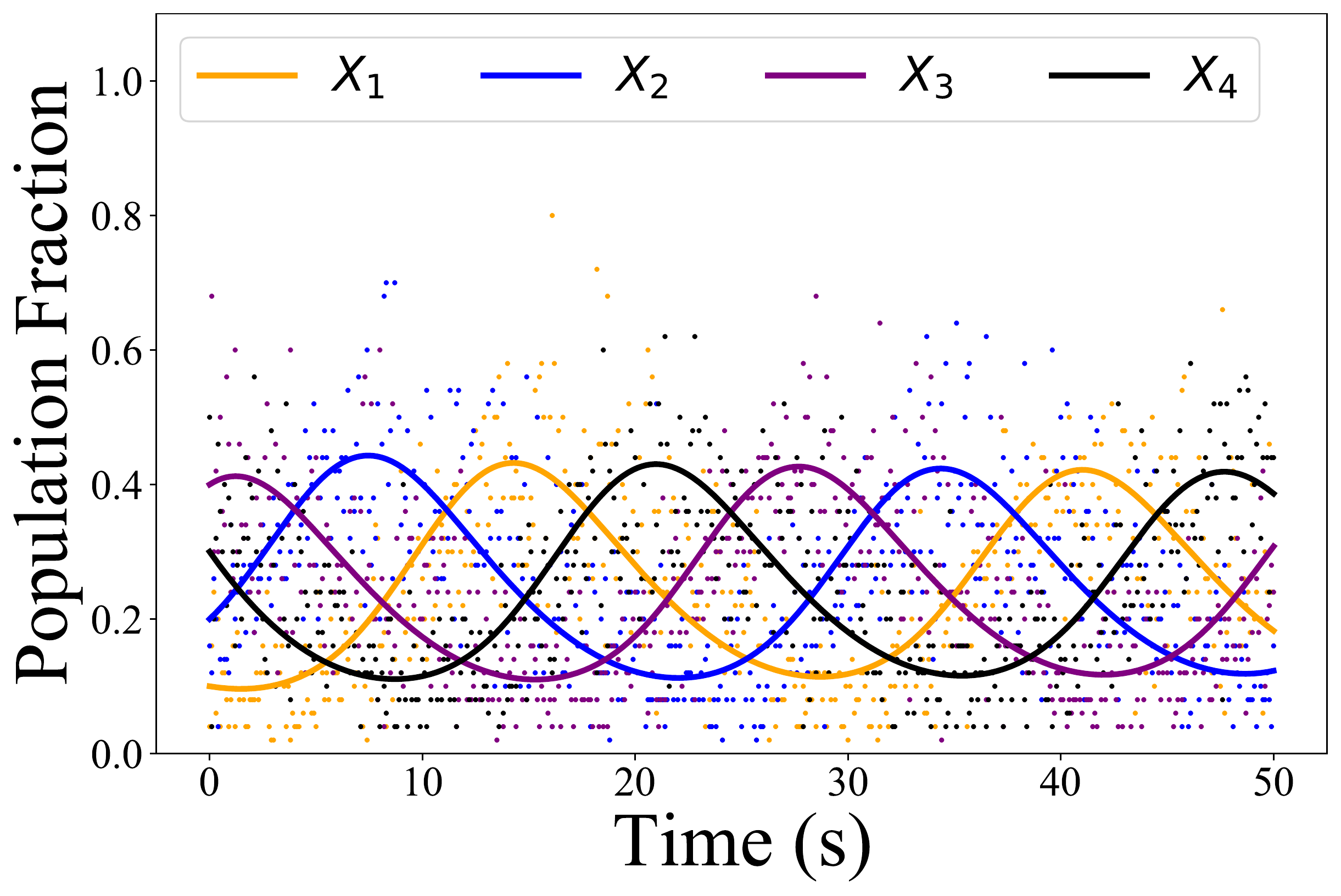}
    \caption{$\mathcal{R}_i = 5$ Microscopic Result}
    \label{fig:example_2_decentralized_micro_r5}
    \end{subfigure}
    \caption{Controlled microscopic results of Example 2 for two distributed estimation cases. Figure (a) uses sensing radius, $\mathcal{R}_i = 10$, and Figure (b) uses sensing radius, $\mathcal{R}_i = 5$. Both trials used $\hat{Y}(0) = \begin{bmatrix} 5 & 10 & 20 & 15 \end{bmatrix}^T$, $\mu = 0.05$, $\alpha_{12} = \alpha_{23} = \alpha_{34} = \alpha_{41} = 0.5$.}
    \label{fig:micro_validation_example_2}
\end{figure}

Microscopic simulation trials are shown in Figure \ref{fig:micro_validation_example_1}, which provides comparisons between the centralized and distributed approach. 
The distributed method shown in Figure \ref{fig:example_1_decentralized_micro} was not as accurate but roughly follows the trajectory which is to be expected. 
In Figure \ref{fig:micro_validation_example_2} a direct comparison between different sensing radii is studied, as $\mathcal{R}_i$ reduces the number of neighbors reduces giving less accurate estimates for the robot task distributions.
The rate that $\mathcal{R}_i$ loses accuracy is related to the density of robots in the environment. 
These microscopic scenarios are the most idealized version of environmental monitoring task assignment.
 
\begin{figure}[ht]
  \centering
    \begin{subfigure}{0.495\textwidth}
    \centering
    \includegraphics[width=\textwidth]{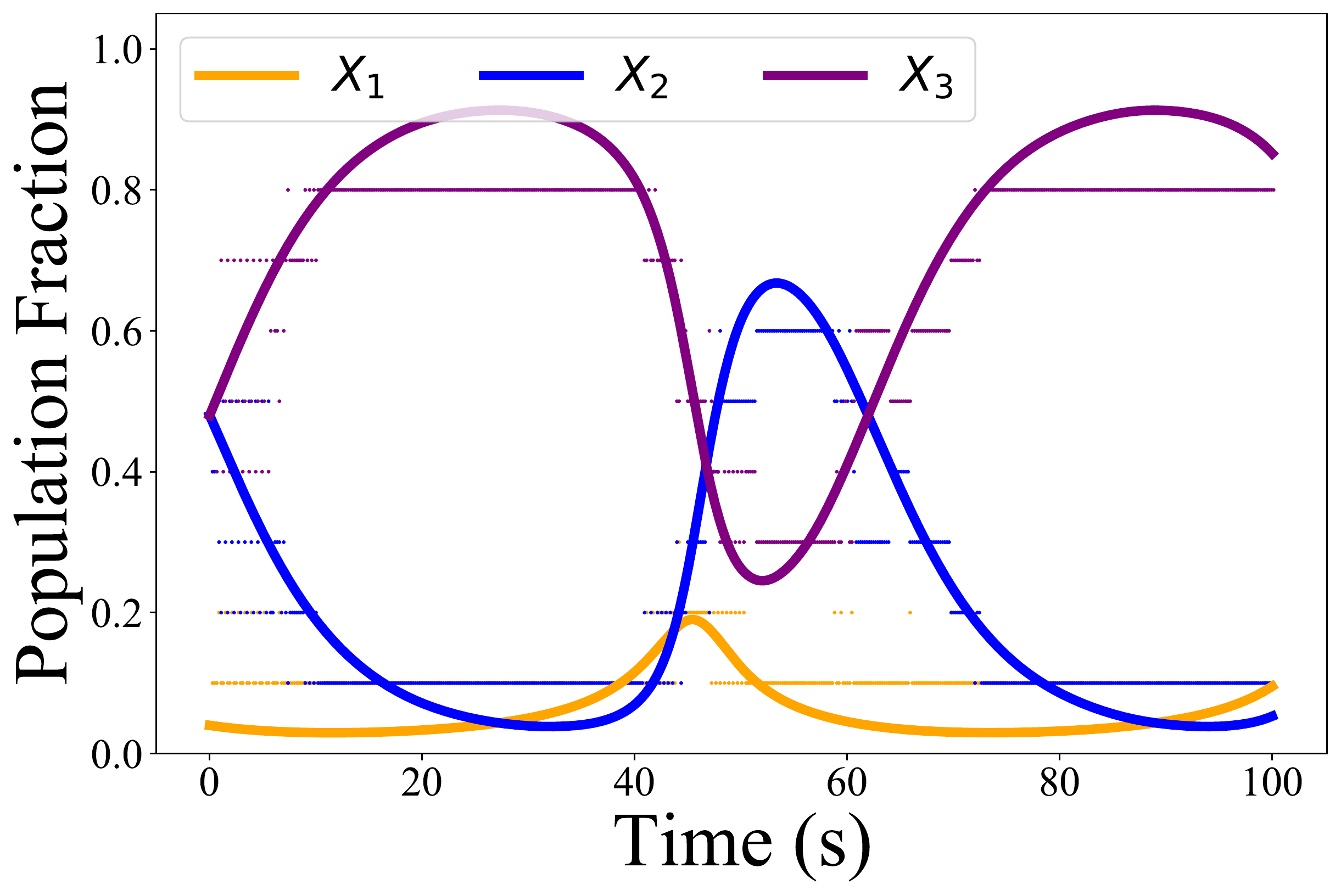}
    \caption{Mixed Reality Centralized Result}
    \label{fig:exp_centralized}
    \end{subfigure}
    \begin{subfigure}{0.495\textwidth}
    \centering
    \includegraphics[width=\textwidth]{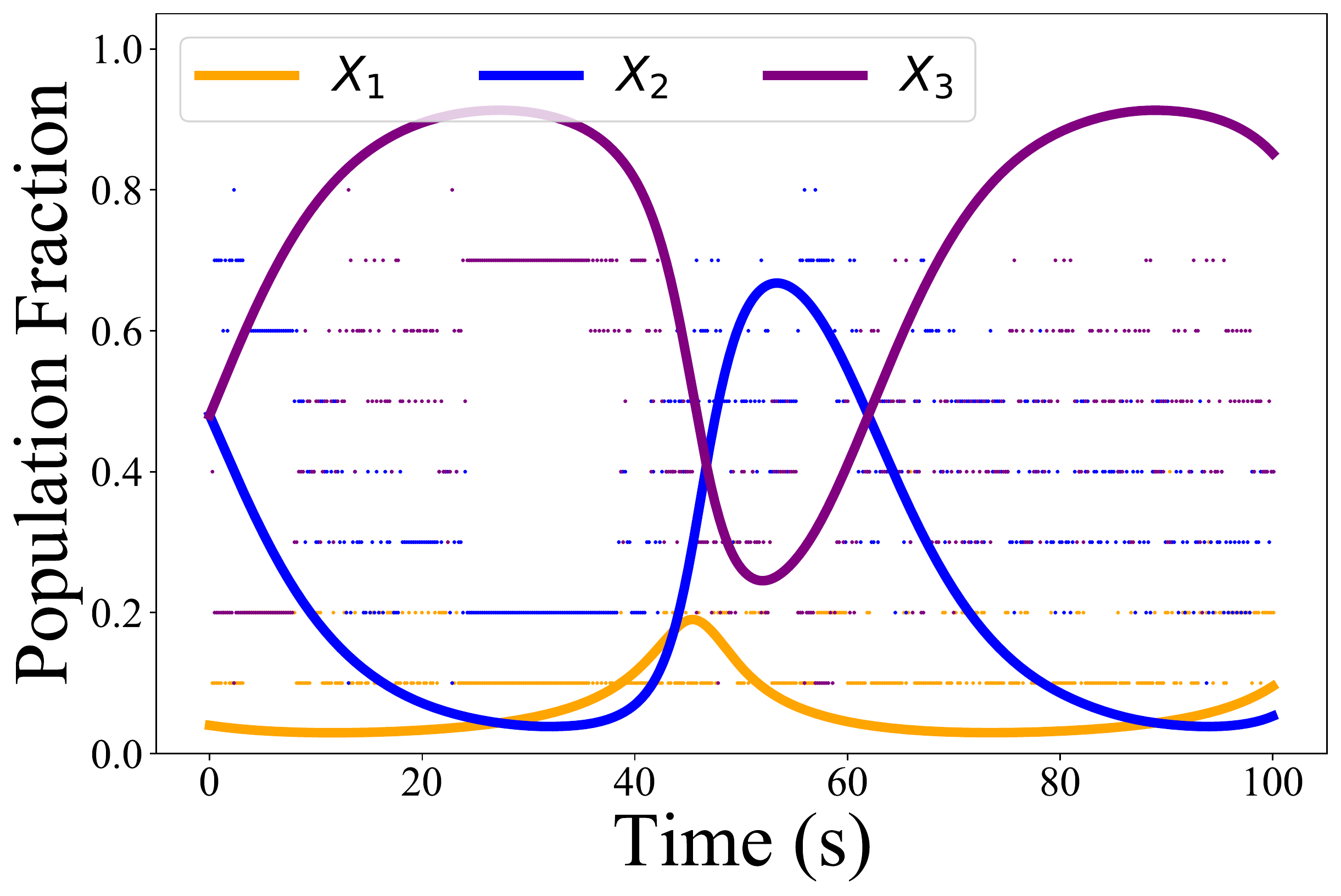}
    \caption{Mixed Reality Distributed Result}
    \label{fig:exp_distributed}
    \end{subfigure}
    \caption{Figure (a) shows a centralized mixed reality experiment, and Figure (b) shows a distributed mixed reality experiment with $\mathcal{R}_i = 0.25$. The parameters used are the same as those used in Figure \ref{fig:micro_validation_example_1}. The environment area is $2.3 \mathrm{m} \times 1.3 \mathrm{m}$ and $\hat{Y}(0) = \begin{bmatrix} 2 & 4 & 4\end{bmatrix}^T$. }
    \label{fig:exp_results}
\end{figure}

Our final evaluation demonstrates that using physical robots in a mixed reality setting produce comparable results.
The mixed reality experimental trials have robots and simulated agents respond to one another in the environment and the desired distributions of behaviors are achieved in Figure \ref{fig:exp_results}.
Notice that in the centralized case, Figure \ref{fig:exp_centralized}, the populations closely follow the desired trajectory.
The variability in the distributed estimation plot, Figure \ref{fig:exp_distributed}, is due to the breakdown of the underlying assumption that within the local sensing radius, $\mathcal{R}_i$, the robot has a decent sample of other robots performing tasks.
Video demonstrations of the mixed reality experiments can be found at \url{https://youtu.be/X0QHgYM1s-g}.

\section{Discussion}

Across multiple types of experimental results, we have demonstrated that we can control time-varying distributions of robots. 
Our results consider the most ideal case, where robots move randomly and consistently throughout the environment with parameters that were selected near equilibria of interest to ensure that our desired behavior was observed. 
There are several directions of future work which need to be addressed to further enhance this approach. 
For example, a more realistic robot behavior should be considered where there may be prolonged stops to complete a task or complex non-random behavior which would change the reaction dynamics of our macroscopic model.
Likewise, right now, we have selected parameters to show that we can achieve time-varying distributions of robots, however, in practice we want the team to evolve based on information sensed in the environment. 
One way to address these nonidealities is to consider feedback from the microscopic team to the macroscopic model.
This would require taking data collected from the individual robot and transforming it into a form that is compatible with the macroscopic model, for example by determining collaboration rate parameters $k_{ij}$. 

A direction of great interest to the authors is to consider more broadly the classes of replicator dynamics in general which achieve our task objective and potentially other more diverse task objectives. 
Important properties of the replicator dynamics which we have identified with our two examples is the need for equilibria to have either limit cycle type behavior or center behavior. 
For example, the resulting oscillations would not exist for parameters or equilibria which resulted in a payoff matrix that had only real eigenvalues, which we know from dynamical systems theory results in stable solutions \cite{guckenheimer2013}. 

Currently, we enforce that at least one agent must be performing a task at any given time to avoid extinction events. 
Extinction can occur when the parameters selected drive the populations to have extreme shifts in size, for example, all agents performing one task. 
Without our stated assumption, there is no way for the team to recover the other tasks if extinction happens. 
A solution would be to consider equilibria which do not cause such extreme population fluctuations. 
Alternatively, there are different nonlinear systems, for example, the Replicator Mutator equations \cite{pais2012hopf} which introduce variability to the model to help ensure that only traits which are strong persist. 

Finally, during the mixed reality experimental trials, robot collisions can occur and are non-catastrophic due to the small size and low speed of the mASVs.
The addition of collision avoidance requires an additional scaling term to compute the interaction radius and is a direction for future work. 
An additional direction of future work is to consider how to expand experimental results to further improve the distributed estimation using robot-robot communication.
A potential approach similar to \cite{Lerman06}, would introduce more stable results by considering a history of observations made about neighbors. 


\section{Conclusion}
In this work, we address the limitations of existing macroscopic modeling techniques to model and control time-varying distributions of robot behaviors. 
This is of particular importance for environmental monitoring tasks, where natural phenomena might require the populations of robots to change behavior, for example, tidal shifts. 
Our proposed methods use robot-robot interactions, which are explicitly ignored by existing linear mean field models.
The result is a rich set of time-varying robot behavior distributions that were not previously achievable. 
In addition, to counteract known uncertainty during robot team deployment, we propose a feedback control strategy which follows a specified desired distribution. 
Our macro-discrete results show that we get improved control over the model, and the idealized microscopic examples support that this is a fruitful avenue for further study. 
Future work includes understanding different types of nonlinear models, and further characterizing the uncertainty by seeking upper and lower bounds on the macroscopic model.  

\section*{Acknowledgements}
We gratefully acknowledge the support of ARL DCIST CRA W911NF-17-2-0181,
Office of Naval Research (ONR)
Award No. N00014-22-1-2157,
and the National Defense Science \& Engineering Graduate (NDSEG) Fellowship Program.

\bibliographystyle{spmpsci}
\bibliography{nonlinear_macro_modeling}

\begin{thebibliography}{10}
\providecommand{\url}[1]{{#1}}
\providecommand{\urlprefix}{URL }
\expandafter\ifx\csname urlstyle\endcsname\relax
  \providecommand{\doi}[1]{DOI~\discretionary{}{}{}#1}\else
  \providecommand{\doi}{DOI~\discretionary{}{}{}\begingroup
  \urlstyle{rm}\Url}\fi

\bibitem{almadhoun2019survey}
Almadhoun, R., Taha, T., Seneviratne, L., Zweiri, Y.: A survey on multi-robot
  coverage path planning for model reconstruction and mapping.
\newblock SN Applied Sciences \textbf{1}(8), 1--24 (2019)

\bibitem{berman2009}
Berman, S., Hal{\'a}sz, A., Hsieh, M.A., Kumar, V.: Optimized stochastic
  policies for task allocation in swarms of robots.
\newblock IEEE transactions on robotics \textbf{25}(4), 927--937 (2009)

\bibitem{biswal2021decentralized}
Biswal, S., Elamvazhuthi, K., Berman, S.: Decentralized control of multi-agent
  systems using local density feedback.
\newblock IEEE Transactions on Automatic Control  (2021)

\bibitem{Deshmukh2018}
Deshmukh, V., Elamvazhuthi, K., Biswal, S., Kakish, Z., Berman, S.: Mean-field
  stabilization of markov chain models for robotic swarms: Computational
  approaches and experimental results.
\newblock IEEE Robotics and Automation Letters \textbf{3}(3), 1985--1992
  (2018).
\newblock \doi{10.1109/LRA.2018.2792696}

\bibitem{dey2018feedback}
Dey, B., Franci, A., {\"O}zcimder, K., Leonard, N.E.: Feedback controlled
  bifurcation of evolutionary dynamics with generalized fitness.
\newblock In: 2018 Annual American Control Conference (ACC), pp. 6049--6054.
  IEEE (2018)

\bibitem{Elamvazhuthi_2019}
Elamvazhuthi, K., Berman, S.: Mean-field models in swarm robotics: a survey.
\newblock Bioinspiration {\&} Biomimetics \textbf{15}(1), 015001 (2019).
\newblock \doi{10.1088/1748-3190/ab49a4}.
\newblock \urlprefix\url{https://doi.org/10.1088/1748-3190/ab49a4}

\bibitem{gerkey2003}
Gerkey, B.P., Mataric, M.J.: Multi-robot task allocation: Analyzing the
  complexity and optimality of key architectures.
\newblock In: 2003 IEEE International Conference on Robotics and Automation
  (ICRA), vol.~3, pp. 3862--3868. IEEE (2003)

\bibitem{gillespie1977exact}
Gillespie, D.T.: Exact stochastic simulation of coupled chemical reactions.
\newblock The journal of physical chemistry \textbf{81}(25), 2340--2361 (1977)

\bibitem{guckenheimer2013}
Guckenheimer, J., Holmes, P.: Nonlinear oscillations, dynamical systems, and
  bifurcations of vector fields, vol.~42.
\newblock Springer Science \& Business Media (2013)

\bibitem{harwell2021characterizing}
Harwell, J., Sylvester, A., Gini, M.: Characterizing the limits of linear
  modeling of non-linear swarm behaviors.
\newblock arXiv preprint arXiv:2110.12307  (2021)

\bibitem{hofbauer1981}
Hofbauer, J.: On the occurrence of limit cycles in the volterra-lotka equation.
\newblock Nonlinear Analysis: Theory, Methods \& Applications \textbf{5}(9),
  1003--1007 (1981)

\bibitem{hsieh2008biologically}
Hsieh, M.A., Hal{\'a}sz, {\'A}., Berman, S., Kumar, V.: Biologically inspired
  redistribution of a swarm of robots among multiple sites.
\newblock Swarm Intelligence \textbf{2}(2), 121--141 (2008)

\bibitem{Hsieh09}
Hsieh, M.A., Halasz, A., Cubuk, E.D., Schoenholz, S., Martinoli, A.:
  Specialization as an optimal strategy under varying external conditions.
\newblock In: 2009 IEEE International Conference on Robotics and Automation
  (ICRA), pp. 1941--1946 (2009).
\newblock \doi{10.1109/ROBOT.2009.5152798}

\bibitem{khamis2015multi}
Khamis, A., Hussein, A., Elmogy, A.: Multi-robot task allocation: A review of
  the state-of-the-art.
\newblock Cooperative Robots and Sensor Networks 2015 pp. 31--51 (2015)

\bibitem{lee2019adaptive}
Lee, W., Kim, D.: Adaptive approach to regulate task distribution in swarm
  robotic systems.
\newblock Swarm and evolutionary computation \textbf{44}, 1108--1118 (2019)

\bibitem{leonard2014multi}
Leonard, N.E.: Multi-agent system dynamics: Bifurcation and behavior of animal
  groups.
\newblock Annual Reviews in Control \textbf{38}(2), 171--183 (2014)

\bibitem{Lerman06}
Lerman, K., Jones, C., Galstyan, A., Matarić, M.J.: Analysis of dynamic task
  allocation in multi-robot systems.
\newblock The International Journal of Robotics Research \textbf{25}(3),
  225--241 (2006).
\newblock \doi{10.1177/0278364906063426}.
\newblock \urlprefix\url{https://doi.org/10.1177/0278364906063426}

\bibitem{Lerman05review}
Lerman, K., Martinoli, A., Galstyan, A.: A review of probabilistic macroscopic
  models for swarm robotic systems.
\newblock In: E.~{\c{S}}ahin, W.M. Spears (eds.) Swarm Robotics, pp. 143--152.
  Springer Berlin Heidelberg, Berlin, Heidelberg (2005)

\bibitem{mather2011}
Mather, T.W., Hsieh, M.A.: Distributed robot ensemble control for deployment to
  multiple sites.
\newblock Robotics: Science and Systems VII  (2011)

\bibitem{Nam17}
Nam, C., Shell, D.A.: Analyzing the sensitivity of the optimal assignment in
  probabilistic multi-robot task allocation.
\newblock IEEE Robotics and Automation Letters \textbf{2}(1), 193--200 (2017).
\newblock \doi{10.1109/LRA.2016.2588138}

\bibitem{pais2012hopf}
Pais, D., Caicedo-Nunez, C.H., Leonard, N.E.: Hopf bifurcations and limit
  cycles in evolutionary network dynamics.
\newblock SIAM Journal on Applied Dynamical Systems \textbf{11}(4), 1754--1784
  (2012)

\bibitem{prorok2017impact}
Prorok, A., Hsieh, M.A., Kumar, V.: The impact of diversity on optimal control
  policies for heterogeneous robot swarms.
\newblock IEEE Transactions on Robotics \textbf{33}(2), 346--358 (2017)

\bibitem{ravichandar2020}
Ravichandar, H., Shaw, K., Chernova, S.: Strata: unified framework for task
  assignments in large teams of heterogeneous agents.
\newblock Autonomous Agents and Multi Agent Systems \textbf{34}(2), 38 (2020)

\bibitem{schuster1981selfregulation}
Schuster, P., Sigmund, K., Hofbauer, J., Gottlieb, R., Merz, P.: Selfregulation
  of behaviour in animal societies.
\newblock Biological Cybernetics \textbf{40}(1), 17--25 (1981)

\bibitem{sigmund1986survey}
Sigmund, K.: A survey of replicator equations.
\newblock In: Complexity, Language, and Life: Mathematical Approaches, pp.
  88--104. Springer (1986)

\end{thebibliography}

\end{document}